\documentclass[runningheads]{llncs}
\usepackage[T1]{fontenc}
\usepackage{graphicx}

\usepackage{tabularx}
\usepackage{comment}
\usepackage{wrapfig}

\begin{document}
\title{Applying Cognitive Design Patterns \\ to General LLM Agents}

\author{Robert E. Wray\orcidID{0000-0002-5311-8593} \and
James R. Kirk\orcidID{0000-0002-2459-3643} \and
John E. Laird\orcidID{0000-0001-7446-3241}}%

\institute{Center for Integrated Cognition, IQMRI, Ann Arbor, MI 28105 USA
\email{\{robert.wray,james.kirk,john.laird\}@cic.iqmri.org}\\
\url{http://integratedcognition.ai}}

\maketitle              %
\begin{abstract}
One goal of AI (and AGI) is to identify and understand specific mechanisms and representations sufficient for general intelligence. Often, this work manifests in research focused on architectures and many cognitive architectures have been explored in AI/AGI. However, different research groups and even different research traditions have somewhat independently identified similar/common patterns of processes and representations or \textit{cognitive design patterns} that are manifest in existing architectures. Today, AI systems exploiting large language models (LLMs) offer a relatively new combination of mechanisms and representations available for exploring the possibilities of general intelligence. This paper outlines a few recurring cognitive design patterns that have appeared in various pre-transformer AI architectures. We then explore how these patterns are evident in systems using LLMs, especially for reasoning and interactive (``agentic'') use cases. By examining and applying these recurring patterns, enables predictions of gaps or deficiencies in today's Agentic LLM Systems and identification of subjects of future research towards general intelligence using  generative foundation models.
\keywords{Agents  \and Cognitive Architecture \and Large Language Models}
\end{abstract}
\section{Introduction}

Cognitive architectures encapsulate and represent theories and commitments toward a general systems architecture for intelligence. More than one hundred different cognitive architectures have been proposed and developed, drawing from many distinct (and often quite disparate) intellectual traditions \cite{kotseruba_40_2020}. Remarkably, as architectures have evolved in response to research outcomes, even though they draw from very different sources, there has been notable convergence and even consensus around both a high-level functional architecture of cognition \cite{laird_standard_2017} as well as many lower-level algorithmic \cite{jones_comparative_2006} and representational \cite{cohen_herbal_2005,crossman_high_2004} commitments. A tentative but hopeful inference from such convergence is that, as a field, we are beginning to understand what components are necessary (or at least important) for realizing artificial general intelligence (AGI).

In contrast to the comparatively long history of cognitive architecture, large language and multi-modal models (LLMs\footnote{Throughout, we use \textit{LLM} as the common term used to describe these models, although most recent models have input and output modalities beyond text/language.}) have emerged in the past five years as a major new technology. The attention is motivated by the breadth and depth of these models and their applicability to many different use cases and applications. While it is sometimes claimed that LLMs alone might offer the potential for AGI \cite{bubeck_sparks_2023}, a new and rapid growing research area is directed toward exploiting LLMs as one (important) component within a larger collection of components and tools that comprise a general AI system \cite{kambhampati_position_2024,zaharia_shift_2024,luo_large_2025}. 

Here, we examine systems architectures comprised primarily of LLM components (in contrast to hybrid integrations of LLMs with 
planners \cite{kambhampati_position_2024}, constraint solvers \cite{lawless_i_2024},
traditional cognitive architectures \cite{laird_proc_2023}, etc.). Such \textit{Agentic LLM Systems} often are developed for specific tasks or domains, such as 
software development \cite{dong_self-collaboration_2024}
or web-content publishing \cite{shao_assisting_2024}. 
However, researchers and developers are also attempting to identify and to develop agentic frameworks sufficiently general to apply to any task \cite{yao_react_2023,shinn_reflexion_2023,wu_autogen_2024}. Further, Agentic LLM Systems are increasingly enhanced with specialized memories also common in cognitive architecture (e.g., episodic memory). In sum, the search for general Agentic LLM Systems shares many goals and non-trivial overlap with cognitive architecture research. 

Within the Agentic LLM community, there have been some high-level proposals to draw lessons from cognitive architecture \cite{sumers_cognitive_2023} and agent architectures and multi-agent systems \cite{sypherd_practical_2024}. In this paper, from the perspective of cognitive- and agent-architecture researchers, we briefly outline the notion of common mechanisms and representations that recur across different architectures. We then review a few of these \textit{cognitive design patterns} and explore their potential relevance to Agentic LLM Systems.  Our examples include both patterns that are currently being explored in Agentic LLMs and ones, relatively underexplored to date, that appear apt for research attention. We contend that such fine-grained, comparative and integrative analysis across divergent fields can both speed future research in Agentic LLMs and also contribute to broadening the scope of research directed toward identifying architectures for general intelligence.

\section{Cognitive Design Patterns}
In software engineering, a design pattern is a description of or template for a solution to a particular design problem. %
Design patterns are typically specified abstractly (not code) and are flexible enough to be adapted/customized for specific situations. \textit{Cognitive design patterns} represent a similar concept. A cognitive design pattern denotes some function/process or representation/memory that routinely appears in agents and/or cognitive architectures. That is, like software design patterns, cognitive design patterns summarize, at an abstract level, common (or at least recurring) solutions to functional requirements for intelligent systems. Table~\ref{tab:cognitive_design_patterns} lists some examples of cognitive design patterns, drawn from past comparative analyses from various researchers \cite{jones_comparative_2006,laird_standard_2017,kotseruba_40_2020,silva_bdi_2020}. For illustration, we include examples from the ACT-R \cite{anderson_integrated_2004} and Soar \cite{laird_soar_2012} cognitive architectures and from the belief-desire-intention (BDI) family of agent architectures \cite{wooldridge_reasoning_2000}. We introduce cognitive design patterns to support the organization and unification of such analyses.\footnote{The coinage here is not new or unique. The the recognition of common patterns occurring across cognitive and agent architectures goes back many years \cite{genesereth_comparative_1991} and has been of long-standing interest to the authors \cite{wray_organizing_1995,jones_comparative_2006,taylor_behavior_2004,crossman_high_2004,laird_standard_2017}.} 

\begin{table}[t]
\caption{Examples of Cognitive Design Patterns.}
\begin{tabular}{p{.33\linewidth}p{.66\linewidth}}
\hline
Cognitive Design Pattern & Examples \\
\hline \hline
Observe-decide-act & BDI: analyze, commit, execute \newline
                    Soar: elaborate/propose, decide, apply (operators) \\ \hline
3-stage \newline memory commitment  & BDI: desire, intention, intention reconsideration \newline Soar: operator proposal, selection, retraction \newline Soar: elaboration, instantiation, JTMS reconsideration \\
\hline 
Hierarchical decomposition & BDI: hierarchical task networks (HTNs) \newline Soar: operator no-change impasses \\ \hline
Short-term (context) \newline memory & ACT-R: buffers (goal, retrieval, visual, manual, ...) \newline Soar: working memory \\ \hline
Ahistorical KR/memory & ACT-R, Soar: semantic memory \\ 
~~~~Retrieval: & ACT-R, Soar: activation-mediated association \\ \hline
Historical KR/memory & Soar: episodic memory \\ %
~~~~Retrieval:  &Soar: cue/memory overlap (encoding specificity) \\ \hline
Procedural KR/memory & ACT-R, Soar: productions, BDI: plans \\ 
~~~~Retrieval: & ACT-R, Soar: associative production-condition match \\ 
~~~~Learning:  & ACT-R, Soar: knowledge compilation/chunking \\
\hline

\end{tabular}
\label{tab:cognitive_design_patterns}
\end{table}

\begin{wrapfigure}[14]{R}{60mm}
    \centering
    \vspace{-2\baselineskip}
    \includegraphics[width=0.99\linewidth]{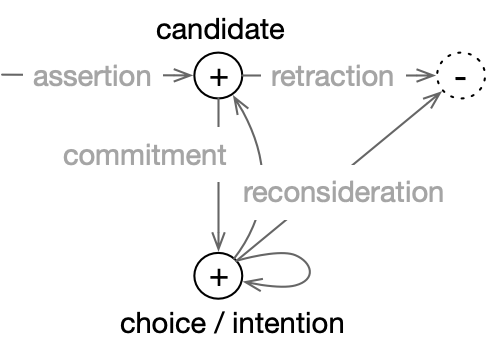}
    \caption{Illustration of the 3-stage commitment cognitive design pattern.}
    \label{fig:3-stage-commitment}
    \vspace{-2\baselineskip}
\end{wrapfigure}

Cognitive design patterns, like software design patterns, are abstract, eliding not only  implementation details, but specification of algorithms. Distinct algorithms and representational commitments are employed to instantiate a cognitive design pattern in different architectures. An is illustrated in Figure~\ref{fig:3-stage-commitment}, many architectures use a 3-stage process to control asserting and retracting individual memory items (rather than the more common store/erase, 2-stage process). Instead of direct assertion, an agent generates candidates for some memory slot. It then uses a selection or \textit{commitment} process \cite{wooldridge_reasoning_2000} to choose among the candidates before making a final choice or \textit{intention}. 

Any commitment can be evaluated or \textit{reconsidered} after the commitment decision \cite{schut_theory_2004}.  Reconsideration can lead to removal/retraction, deselection (in which a previous choice is demoted to a candidate), or reaffirmation (continuing with the current choice). Both BDI architectures and Soar (and some additional cognitive architectures) exhibit this 3-stage process in their primary unit of deliberation. However, the representational foci of deliberation (plans in BDI, operators in Soar) are distinct, as are the specific algorithms used for assertion, commitment, and reconsideration \cite{wray_architectural_2003,schut_theory_2004,jones_comparative_2006}.

Familiar algorithms may be specialized (in scope and/or implementation) to realize a cognitive function within an architecture. For example, Soar uses a justification-based truth maintenance system (JTMS) \cite{doyle_truth_1979} as the reconsideration process for one functional type of memory element. The JTMS implementation is specialized for its integration within Soar. Cognitive design patterns are useful as a unit of comparison and analysis because abstracting to just the functional role (reconsideration) distances the analysis from implementation details (a customized JTMS for Soar), making more evident and clear commonalities and contrasts in architectural commitments. If we just assert that Soar uses a JTMS and a BDI architecture uses some decision-theoretic calculations over different types of memory instances, it is easy to miss that the functional role of both of these processes, reconsideration (and more broadly, belief revision and context/memory management) is comparable in both architectures.

\section{Cognitive Design Patterns \& Agentic LLM Systems}
Cognitive design patterns appear and recur in many existing systems motivated toward achieving general intelligence. We now consider if/how these cognitive design patterns might be apt for understanding, evaluating, and anticipating the course of Agentic LLM research. We consider the following three questions:

\begin{enumerate}
    \item What cognitive design patterns are evident in existing Agentic LLM systems? If cognitive design patterns represent a catalog of functions important to general intelligence, we should observe them in Agentic LLM systems targeted toward general intelligence. Some cognitive design patterns are clearly evident. We discuss two recent examples in some detail.
    \item What cognitive design patterns appear apt for Agentic LLMs but are not yet part of the Agentic LLM mainstream? We present two examples that have not yet been deeply investigated and that appear relevant in the current state-of-the-art in Agentic LLMs systems.
    \item Do the unique properties of Agentic LLMs suggest new cognitive design patterns for general intelligence? We examine how computational and behavioral characteristics of Agentic LLMs may lead to new design patterns.
\end{enumerate}

We now consider each of these questions, drawing examples from research in both Agentic LLMs and cognitive and agent architectures. We identify specific examples to illustrate and provide insight. An exhaustive survey for each of these questions is too broad for a conference paper and, given the accelerating pace of exploration of these topics, somewhat impractical.

\begin{table}[tbh]
\caption{Cognitive design patterns evident in recent Agentic LLM research.}
\begin{tabular}{p{.3\linewidth}p{.69\linewidth}}
\hline
Cognitive Design Pattern & Agent LLM Systems \\
\hline \hline
Observe-decide-act &  ReAct \cite{yao_react_2023}, Reflexion \cite{shinn_reflexion_2023} \\ \hline
Hierarchical \newline decomposition & Voyager \cite{wang_voyager_2024}, ADaPT \cite{prasad_adapt_2024}, DeAR \cite{xue_decompose_2024}, \newline Tree of Thoughts \cite{yao_tree_2023}, Graph of Thoughts \cite{besta_graph_2024}\\ \hline
Knowledge compilation & NL: ExpeL \cite{zhao_expel_2024}, Reflexion \cite{shinn_reflexion_2023}  \newline
Structured Reps: Voyager \cite{wang_voyager_2024}, LLMRG \cite{wang_llmrg_2024}, \newline
\quad\null~~~~~~~~~~~~~~~~~~~~~~~~cognitive agent framework \cite {zhu_bootstrapping_2024} \\ \hline

Context memory & MemGPT \cite{packer_memgpt_2024}, LONGMEM \cite{wang_augmenting_2023}  \\ \hline 
Ahistorical KR/memory &  A-MEM \cite{xu_-mem_2025}, WISE \cite{wang_wise_2024}, MemoryBank \cite{zhong_memorybank_2024}\\ \hline
Historical KR/memory &  Generative Agents \cite{park_generative_2023}, MemoryBank \cite{zhong_memorybank_2024} \\ \hline
Procedural KR/memory & ProgPROMPT \cite{singh_progprompt_2023}, Voyager \cite{wang_voyager_2024}, Self-evolving GPT \cite{gao_self-evolving_2024}   \\ \hline

\end{tabular}
\label{tab:evident_patterns}
\end{table}

\subsection{Cognitive Design Patterns in Existing Systems}
Many of the cognitive design patterns in Table~\ref{tab:cognitive_design_patterns} are well-known, and examples are increasingly evident in Agentic LLM research. Table~\ref{tab:evident_patterns} lists examples from recent research.\footnote{Because our goal is to illustrate how prevalent the exploration of AGI cognitive design patterns is in Agentic LLM systems, the list is purposely inclusive. The table does not distinguish full vs. partial realizations of cognitive design patterns.} We highlight two specific examples.

\subsubsection{Observe-Decide-Act in ReAct}

\begin{figure}[tb]
    \centering
    \includegraphics[width=0.99\linewidth]{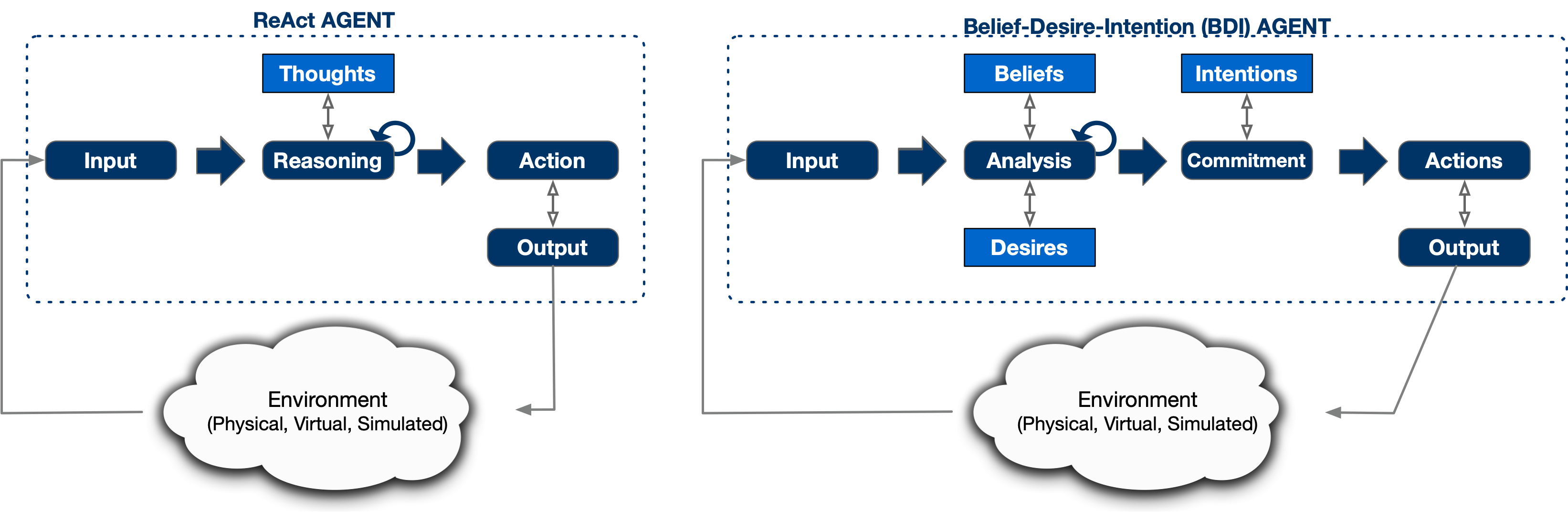}
    \caption{Contrasting Decision/Action cycles in the ReAct (left) and BDI (right) paradigms. BDI illustrates the Observe-Decide-Act cognitive design pattern.}
    \label{fig:bdi-ooda}
\end{figure}

Generally, LLM performance improves when the user prompt directs the model to be explicit about its reasoning, as in the now ubiquitous Chain-of-Thought (CoT) prompting method \cite{wei_chain--thought_2022}.  ReAct \cite{yao_react_2023}, one of the earliest and most influential Agentic LLM methods, builds on CoT by distinguishing between internal and external problem-solving steps. As illustrated in Figure~\ref{fig:bdi-ooda}~(left), ReAct initially prompts the LLM for natural-language (NL) statements (\textit{thoughts}) about the problem before taking any action.  While subsequent research has interrogated the relative contributions of examples vs. decomposition in ReAct    \cite{bhambri_think_2025}, it did demonstrate significant improvements in LLM responses across multiple domains \cite{yao_react_2023}. ReAct is now a foundational approach for Agentic LLM systems and has been incorporated into agentic software development tools such as LangGraph \cite{langchain_langgraph_2025}.

ReAct replicates a subset of the common observe-decide-act pattern. Figure~\ref{fig:bdi-ooda}~(right) illustrates how observe-decide-act is frequently implemented in BDI architectures: new input triggers an analysis/reasoning process, informed by the agent's current beliefs and desires. The updated context of beliefs and desires then informs a decision or commitment process (including reconsideration, as discussed previously). The intention(s) that result from commitment then lead to agent actions. The basic decision process of other architectures, such as Soar and ACT-R, can also be mapped to this abstract observe-decide-act pattern. 

ReAct has a significant overlap with this pattern but lacks the step that explicitly makes commitments. Thus, this mapping of ReAct to the familiar design pattern leads to an immediate, empirical question: Would introducing commitment in ReAct result in better overall reasoning outcomes than ReAct alone? We contend that one of the advantages of mapping Agentic LLM solutions to cognitive design patterns is identifying specific questions such as this one.

\subsubsection{Episodic Memory in Generative Agents}
Agentic LLM researchers are increasingly exploring various \textit{types} of long-term memories (LTMs) \cite{wu_human_2025}. Examples include both general-purpose LTMs \cite{zhong_memorybank_2024} and LTMs specialized for  
procedures (or skills) \cite{wang_voyager_2024,gao_self-evolving_2024},
history/episodes \cite{park_generative_2023,pink_position_2025,zhong_memorybank_2024},
or facts (semantic memory) \cite{xu_-mem_2025,wang_scienceworld_2022}. Generally, long-term memories give an Agentic LLM system more ability to control the context it uses for its generative steps, enabling adaptive responses to a dynamic environment over longer periods of time \cite{gao_self-evolving_2024,wu_llms_2025}.

An early, influential example of LTM in Agentic LLM Systems is Generative Agents \cite{park_generative_2023}. Generative Agents simulates human behavior via a collection of LLM agents, each representing a person living in a small village. Agents interact with one another through natural language and undertake various individual and collective tasks, such as attending work or school. Generative Agents implements an LTM that concatenates a history of specific recent experiences (\textit{observations}) and abstract, integrative summaries of past experience (\textit{reflections}). Reflections capture gists of experience (``Klaus worked in the lab all day'') and also insights (``Klaus' dedication to his lab work suggests he is passionate about research.''). 

These memories then inform subsequent agent behavior. Agents retrieve a combination of observations and reflections into a \textit{Memory Stream}, which provides context for agent planning and action. Retrieval uses recency, agent-judged importance, and an agent's subjective sense of relevance to retrieve items from long-term memory into the memory stream.

\begin{table}[tbh]
\caption{Characteristics of Episodic Memory \& Comparison to Generative Agents.}
\begin{tabularx}{\linewidth}{p{1in}cX}
\hline
Characteristic & In \cite{park_generative_2023}? & Comments \\
\hline \hline

Learning Process  \\ \hline
Automatic & Semi & A periodic process was used to evaluate memories and save a subset. \\
Autobiographical & Yes & Memories are recorded in the 3rd-person, but agents understand memories as being about themselves. \\
Autonoetic & ? & Unclear how agents distinguish current understanding from memory of past experience. \\ 
Episodic \newline ~segmentation & No & Observation span and reflection prompts are based on pre-defined, static periods (e.g., every 100 observations triggers a reflection).\\ 
Variable Length & Yes & Reflections can span variable lengths of agent experience.\\ \hline

Retrieval Process \\ \hline 
Cue-based  & Partial & Objects present in agent situation used as retrieval cues\\
Spontaneous? & Yes & Retrieval is implemented as a recurring, automated process\\
Deliberate? & No & Agents cannot deliberately attempt to construct cues or retrieve memories\\
Encoding \newline specificity & Partial & Relevance (one of three retrieval criteria) uses semantic similarity, not encoding specificity.
\\ \hline

\end{tabularx}
\label{tab:episodic_memory}
\end{table}

The developers of Generative Agents do not describe its long-term memory as an episodic memory but it is often cited as an example of episodic memory using LLMs. Episodic memory is autobiographical,  memory of what happened to me. Specific characteristics of human episodic memory are well-understood \cite{tulving_elements_1983} and have been  implemented within cognitive architectures  \cite{nuxoll_enhancing_2012}. 

Comparing any implementation of an episodic memory to the general characteristics of the episodic memory cognitive design pattern can potentially identify capabilities, limitations, and opportunities for further exploration. Table~\ref{tab:episodic_memory} lists specific requirements for computational implementations of episodic memory, adapted from Nuxoll and Laird \cite{nuxoll_enhancing_2012}. The table also indicates which characteristics of episodic memory Generative Agents satisfy (and which it does not). 

To  highlight one example, most conceptions of episodic memory assume the \textit{encoding specificity principle} \cite{tulving_encoding_1973}, which (roughly) says that episodic memories encode event context. Retrieval of some past episode is more likely when retrieval cues match the encoded context: entering a home that one lived in many years ago might trigger memories of being in that home. The context (e.g., the floor plan) was encoded in memory. Being present in the home again creates perceptual cues that match encoded cues from past memories, resulting in retrieval.

Memory in Generative Agents encodes objects, providing some specificity, but the encoding of observations and reflections is not fully contextual. Retrieval employs relevance, which relies on semantic similarity rather than cue-specificity. The table identifies other similarities and differences. As for ReAct, this analysis points to specific research questions, such as exploring how agent behavior might change if the agent could deliberately construct explicit retrieval cues or whether the quality of reflection would change with a more context-sensitive trigger.

\subsection{Patterns Apt for Exploration and Exploitation}

In this section, we briefly discuss two cognitive design patterns that have not yet been taken up in the Agentic LLM mainstream: reconsideration and knowledge compilation. Both patterns appear to be relevant to addressing current shortcomings in Agentic LLMs and thus represent research opportunities.

\subsubsection{Commitment and Reconsideration in Agentic LLMs}
Reconsideration is a cognitive design pattern that describes the process of evaluating whether a prior commitment, such as a goal or a plan, should continue in an agent's current situation. Reconsideration that results in a change in commitment is a non-monotonic step. 
Chat-focused LLMs tend to resist major redirection in the trajectory of their token generation and LLMs alone are not consisent or reliable in producing non-monotonic reasoning steps \cite{leidinger_are_2024}. 

In addition to reflection strategies (discussed below),  reconsideration for intentions or commitments in Agentic LLMs could allow an the agent to periodically assess the continuing viability and desirability of its goals. A few approaches have been developed, often comparable to short-term memories, to limit or to manage what context is presented with each LLM query \cite{packer_memgpt_2024}, comparable to paging in operating systems. This approach does support (if somewhat indirectly) redirection by forgetting/deleting past context. Reconsideration via LLMs would  enhance recent efforts in using Agent LLMs for planning and plan decomposition, which share some similarities to commitment processes \cite{prasad_adapt_2024,xue_decompose_2024}. As LLM Agents make deliberate, explicit commitments, they will then also need to decide if/when those commitments should be abandoned, just as traditional agents do.

\subsubsection{Knowledge Compilation} Multi-step reasoning (or  problem search \cite{laird_universal_1983}) can be computationally expensive, even NP-complete \cite{bylander_complexity_1991}. Knowledge compilation caches the results of reasoning steps into a more compact representation, essentially amortizing the cost of expensive reasoning by enabling immediate use of the cached reasoning steps in future, similar situations. Examples of knowledge compilation abound,  including macro-operator learning in STRIPS \cite{fikes_learning_1972}, various forms of explanation-based learning \cite{mitchell_explanation-based_1986} and implementations in both ACT-R (production compilation \cite{taatgen_production_2003}) and Soar (chunking \cite{laird_chunking_1986}). 

Ignoring quality and reliability issues, %
Agentic LLMs engage in more potentially expensive problem search (i.e., individual query-response steps) that requires more run-time compute than LLMs alone. For example, we reviewed above how ReAct divides internal and external problem-solving steps into separate query-response interactions, thus (at least) doubling the number of query-response interactions. Further, large reasoning models (LRMs) have recently emerged, such as OpenAI o1 and DeepSeek R1, in which fine-grained, multi-step reasoning, similar to Chain-of-Thought, is reinforced in token generation \cite{xu_towards_2025,openai_openai_2024,deepseek-ai_deepseek-r1_2025}. The computational demands of run-time inference in these new reasoning models is significantly greater than chat-based models. An open challenge for LRMs is to reduce their run-time inference cost \cite{xu_towards_2025}.

Given these trends, the knowledge compilation design pattern appears directly relevant to Agentic LLMs and LRMs. Compiling LLM reasoning into a different form for subsequent use has been an active area of research, as a special case of program synthesis. For example, Voyager \cite{wang_voyager_2024} uses LLM reasoning to construct Minecraft crafting recipes (as executable JavaScript), LLMRG \cite{wang_llmrg_2024} constructs structured graphs that summarize user history and preferences in a recommendation system, and Zhu and Reid \cite{zhu_bootstrapping_2024} describe generating Soar productions from LLM reasoning.

While these examples illustrate the value of compiling reasoning into a different form, direct, online caching of LLM responses in natural language has not yet become widely researched.   Voyager generates an NL summary of the functions (skills) that it produces. Reflexion \cite{shinn_reflexion_2023}, a system designed to improve its problem-solving policy via LLM-mediated reflection, caches its reflections in a memory to be used in subsequent trials. This caching supports within-task transfer but offers little/no across-task transfer. ExpeL \cite{zhao_expel_2024}  develops NL \textit{insights} from analysis of initial, exploratory problem-solving traces in a domain. The functionality of ExpeL's insights comes closest (of those we have found) to the knowledge compilation design pattern. However, insight generation requires an explicit, training stage, in sharp contrast with the online use of knowledge compilation for problem solving in ACT-R and Soar.

Finally, knowledge compilation assumes an agent will encounter similar situations in the future, so that saving the results of potentially expensive, multi-step reasoning is worthwhile. 
This assumption is reasonable as long as memory size and retrieval from it are relatively inexpensive \cite{minton_quantitative_1990}. Mitigating this \textit{utility problem} has strongly influenced the implementation of knowledge compilation within cognitive architectures \cite{minton_quantitative_1990,tambe_problem_1990}. As knowledge compilation becomes more commonplace in Agentic LLM systems, we can expect that utility problems will arise and that specialized techniques and algorithms will be needed and researched to enable routine use of knowledge compilation in LLM systems.

\subsection{Emergence of Novel Cognitive Design Patterns}

Most cognitive architectures prescribe a fixed, pre-defined control flow. In contrast, LLM computation intermixes content and control. Given this difference, we can expect novel patterns of processing will likely emerge from integrating and combining LLM modules for Agentic AI.

As illustration, a candidate novel pattern is \textit{step-wise reflection}. Step-wise reflection is intended to improve overall reliability of LLM responses. The simplest versions of step-wise reflection prompt the language model to evaluate the previous response. Self-consistency \cite{wang_self-consistency_2023}, a more sophisticated instance of the pattern, generates multiple LLM responses to an individual prompt. Then, via a reflection step, it identifies the most frequently-occurring (or \textit{consistent}) response and chooses it as a ``final'' answer to the original query.  

While step-wise reflection is functionally similar to reconsideration and belief revision \cite{gardenfors_knowledge_1988}, its specific characteristics are unique. For example, the reflection step itself is as potentially unreliable as the initial response and thus possibly recursive and unbounded in the worst case. It contrasts with Self-reflection \cite{renze_self-reflection_2024} in which an external process provides feedback that a previous response (typically, an answer to a question) is incorrect. In step-wise reflection, a LLM is asked to evaluate correctness and adequacy without external/oracular feedback. 

Finally, step-wise reflection differs from more familiar metacognitive reflection, such as Soar's retrospection \cite{mohan_learning_2014}  or Reflexion \cite{shinn_reflexion_2023}, an extension of ReAct outlined previously. In metacognitive reflection, the agent reflects over many steps of problem solving, including interaction with the environment, to gain additional knowledge (such as generalization or correction). In summary, step-wise reflection appears to be a novel pattern that integrates and combines aspects of other patterns in a unique way, driven by the need for continual and fine-grained (step-wise) assessment of LLM-driven reasoning.

\section{Conclusions}

We presented cognitive design patterns as a powerful analytic tool for organizing and understanding the explosion of research in Agentic LLM systems. Cognitive design patterns encapsulate high-level descriptions of processes, representations, and memories that have recurred in research and development of  agent/cognitive architectures motivated by the goal of AGI. Using these patterns to analyze Agentic LLM systems, we identified potential limitations in existing approaches and  generated predictions that specific extensions/elaboration of the approaches, informed by cognitive design patterns, will mitigate these limitations. 

Long-term, we see cognitive design patterns as a complement to other approaches that are attempting to codify commonalities in approaches to general intelligence, such as the Common Model of Cognition \cite{laird_standard_2017}. Cognitive design patterns represent a functionally motivated approach to organizing cognitive functions, independent of whether those patterns are representative of or relevant to human cognition. Together, we anticipate that such approaches could represent an emerging discipline of \textit{Comparative Cognitive Architecture}. Analysis, understanding, and building on progress across architectural paradigms has been a barrier to faster progress in AGI. We believe an analytic, comparative, and integrative methodology, as a companion to existing architectural implementation research,  has the potential to speed the exploration and progress toward general, artificially intelligent agents.

\begin{credits}
\subsubsection{\ackname} This work was partially supported by the Office of Naval Research, contract N00014-21-1-2369, and the Defense Advanced Research Projects Agency, contract HQ072716D0006. The views and conclusions contained in this document are those of the authors and should not be interpreted as representing the official policies, either expressed or implied, of the Department of Defense, Office of Naval Research, or DARPA. The U.S. Government is authorized to reproduce and distribute reprints for Government purposes notwithstanding any copyright notation hereon. Cognitive design patterns have been a decades-long interest of the authors and developed over many interactions and conversations with our colleagues including Ron Chong, Jacob Crossman, Kevin Gluck, Scott Huffman, Randy Jones, Rick Lewis, Christian Lebiere, Doug Pearson, Frank Ritter, Glenn Taylor, Mike van Lent, and Richard Young. We would like to thank the anonymous reviewers for constructive suggestions and feedback. 

\subsubsection{\discintname}
The authors have no competing interests to declare that are
relevant to the content of this article. 
\end{credits}

\bibliographystyle{splncs04}

\end{document}